\documentclass{article}

\usepackage[accepted]{icml2019}

\usepackage{times}
\usepackage{latexsym}
\usepackage{amsfonts}
\usepackage{graphicx}
\usepackage{dblfloatfix}
\usepackage{multirow}
\usepackage{subcaption}
\graphicspath{ {Figs/} }

\usepackage{url}
\usepackage{hyperref}

\icmltitlerunning{Bootstrapping NLU Models with Multi-task Learning}

\begin{document}

\twocolumn[
\icmltitle{Bootstrapping NLU Models with Multi-task Learning}




\begin{icmlauthorlist}
\icmlauthor{Shubham Kapoor}{goo,intern}
\icmlauthor{Caglar Tirkaz}{to}
\end{icmlauthorlist}

\icmlaffiliation{to}{Amazon Development Center, Aachen, Germany}
\icmlaffiliation{goo}{Amazon Research, Berlin, Germany}
\icmlaffiliation{intern}{Work done during internship at the Amazon Development Center, Aachen, Germany}

\icmlcorrespondingauthor{Shubham Kapoor}{kapooshu@amazon.com}
\icmlcorrespondingauthor{Caglar Tirkaz}{caglart@amazon.com}


\vskip 0.3in
]

\printAffiliationsAndNotice{} 

\begin{abstract}
  Bootstrapping natural language understanding (NLU) systems with minimal training data is a fundamental challenge of extending digital assistants like Alexa and Siri to a new language.
  A common approach that is adapted in digital assistants when responding to a user query is to process the input in a pipeline manner where the first task is to predict the domain, followed by the inference of intent and slots.
  However, this cascaded approach instigates error propagation and prevents information sharing among these tasks.
  Further, the use of words as the atomic units of meaning as done in many studies might lead to coverage problems for morphologically rich languages such as German and French when data is limited.
  We address these issues by introducing a character-level unified neural architecture for joint modeling of the domain, intent, and slot classification.
  We compose word-embeddings from characters and jointly optimize all classification tasks via multi-task learning.
  In our results, we show that the proposed architecture is an optimal choice for bootstrapping NLU systems in low-resource settings thus saving time, cost and human effort.
\end{abstract}

\section{Introduction}

Digital assistants like Amazon Alexa help users with their daily lives for various tasks such as setting up an alarm, booking a taxi, adding events to their calendar or making a dinner reservation.
Since these digital assistants support only a limited set of languages, a rapid expansion of these systems to new languages is a prioritized goal for companies like Amazon, Google and Apple to expand their user base.
However, the task of language expansion is not trivial since it requires large amounts of annotated training data which translates to additional costs, effort and time. 
Hence, developing efficient techniques for training spoken language understanding (SLU) systems in low-resource settings is an active research topic. One of the open challenges to accelerate the pace of language expansion is to bootstrap an accurate natural language understanding (NLU) module for new languages with minimal training data. The NLU module, which is our main focus, is a crucial component of an SLU system and is responsible for deriving the semantic interpretation of a spoken utterance or query. 

The problem of obtaining data for low-resource languages is further amplified when bootstrapping a morphologically rich language such as German, French, Turkish, Hungarian, etc. Such languages can have extensive vocabularies as various forms of the same word can be generated through inflectional and derivational suffixation or compounding. Hence, this results in high lexical sparseness and leads to significant coverage issues for parametric models if words are modeled as the smallest units of meaning.

Apart from the scarcity of annotated training data, another problem is how the NLU modules are designed. The NLU module is generally designed using a pipeline approach \cite{sarikaya2016overview, gupta2006t} where a standard way of interpreting an input utterance starts with predicting its domain followed by domain-specific intent and slots. 
Since the NLU module of digital assistants like Alexa processes multiple domains, such a cascaded model prevents learning any domain-invariant features. Moreover, this pipeline approach in the NLU component propagates the errors from an upstream to a downstream task, which degrades the system's overall performance.
Further, the pipeline approach also prohibits any knowledge sharing between the closely related subtasks of domain, intent and slot classification. 
Such knowledge sharing among the NLU tasks has the potential to enhance the performance of each task in a low-resource setting, thereby boosting the overall performance of the NLU module. 

In this work, we present an end-to-end unified neural architecture for bootstrapping NLU models in low-resource settings for domain, intent and slot classification. The task involves classifying utterances such as 
'\textit{play frozen from madonna}'
and classifying the domain, intent and slots as 
'\textit{Music}', 
'\textit{Play Music}' and 
'\textit{Other Songname Other ArtistName}', respectively.
Firstly, our approach uses character-level modeling to deal with the extensive vocabularies of morphologically-rich languages. This helps with the language coverage issues by efficiently modeling the out of vocabulary (OOV) words encountered during inference phase. Further, this facilitates implicit parameter sharing between the various words with similar subword units, like rain and raining, thus inducing a robust representation of words in low-resource settings. 
Secondly, our approach uses multi-task learning to enable a unified architecture, which jointly infers the output of all of the subtasks and overcomes the problem of error propagation. Since joint optimization facilitates information sharing, our architecture is able to achieve notable accuracy improvements across all NLU tasks. 
Thirdly, we present the analysis of using pre-trained word embeddings when initializing our model to further improve the model performance in low-resource settings. 
Finally, we evaluate the proposed architecture on real world data in a large scale setting and provide detailed analysis of the design choices.

\section{Related Work}

Various architectures have been proposed for joint modelling of the intent and slot classification to reduce the error propagation within a particular domain. Probabilistic models such as a triangular conditional random field (CRF) \cite{jeong2008triangular} and a convolutional neural network based CRF \cite{xu2013convolutional} reports significant performance gains by using a unified model. Since the intent and slots of an utterance are highly correlated, \citet{zhang2016joint} propose a recurrent neural network (RNN) based architecture optimized by both tasks together which operate on a shared embedding. Another RNN-based model \cite{liu2016attention} used an auto-regressive decoder and attention mechanism to learn these both tasks together. An approach that is most similar to ours is proposed by \citet{kim2017onenet} where the authors jointly optimized domain, intent and slot classification tasks. Compared to their work, we enable direct information flow from an upstream to its downstream task and show that this improves the model performance.

In addition to jointly modeling intent and slot task, there exists prior work on domain adaptation focusing on sharing features among multiple-domains.
\citet{jaech2016domain} models multiple domain-specific slot-filling layers together with their roots in a single RNN-based encoder thus jointly training the encoding layers. 
\citet{kim2016frustratingly} presents another approach which uses $K+1$ slot-filling layers where in addition to the domain-specific layer for each of the $K$ domains there exists an additional layer, shared by all domains, for inducing feature augmentation. 

Distributed word embeddings have improved the performance of many NLP tasks like sentiment analysis \cite{maas2011learning}, language modelling \cite{bengio2003neural} and named entity recognition (NER) \cite{turian2010word}. Most of these existing approaches treat words as individual atomic units, thus completely ignoring their internal structure. Furthermore, the quality of word-based embedding models deteriorate for rare and unseen words \cite{bojanowski2017enriching} since some words occur so rarely that there might not be enough instances of a word to learn its representation. Recently, compositional word embeddings have been applied to a variety of NLP tasks like language modelling \cite{vania2017characters}, NER \cite{Guillaume2016neural} and neural machine translation (NMT) \cite{ataman2018compositional} and achieved successful results. Further, it has been shown that morphological and semantic information can be exploited by composing the representations of subwords at different granularity level i.e. characters \cite{kim2016character}, character n-grams \cite{bojanowski2017enriching} and bytes \cite{gillick2016multilingual}.

\begin{figure*}[!t]
  \centering
  \includegraphics[width=0.9\textwidth]{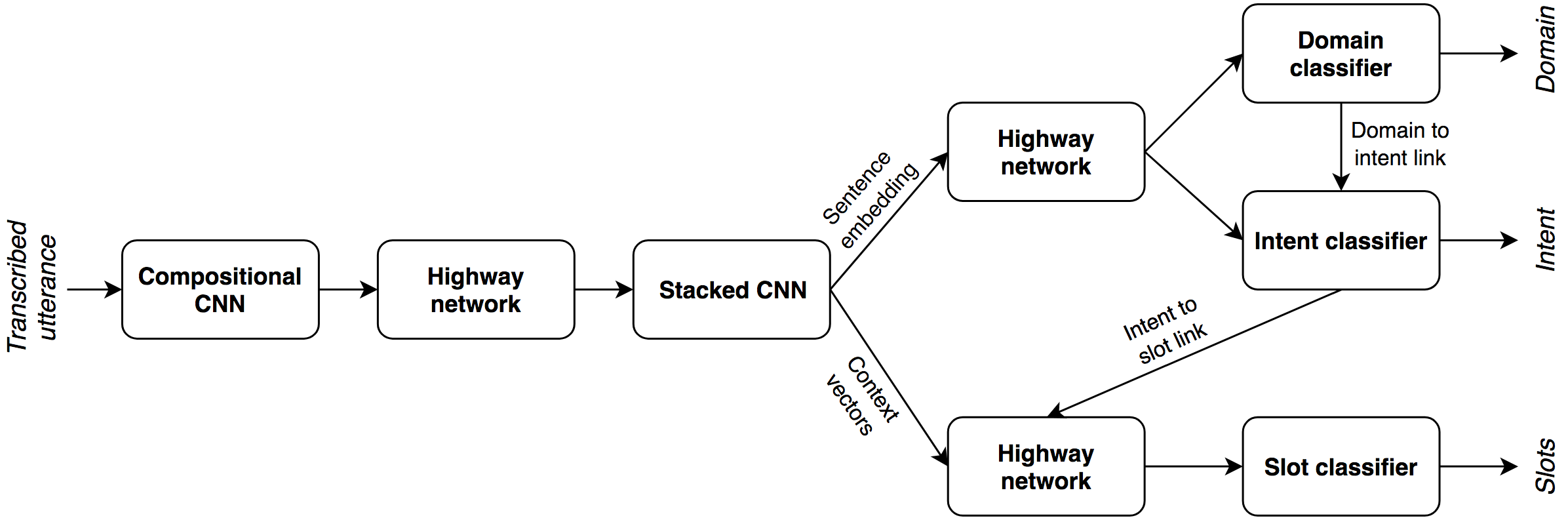}
  \caption{An overview of the proposed end-to-end joint NLU model. 
  The compositional CNN is used to compose word embeddings from characters; 
  highway networks not only facilitate information flow but also enable non linear transformations of the input; 
  multiple convolutional layers in the stacked CNN increase the receptive field when generating context vectors;
  domain-intent and intent-slot links enable information flow from an upstream task to a downstream one and create a way for a downstream task to provide feedback to an upstream task.
  }
  \label{fig:model_arch}   
\end{figure*}

\section{Architecture}

As shown in Fig.~\ref{fig:model_arch}, our model is composed of five major building blocks: (a) the compositional CNN layer (Sec.~\ref{sec:compositional_cnn}) that derives word features from character n-grams, (b) highway layers (Sec.~\ref{sec:highhway_network}) that model interactions among its inputs and facilitate information flow, (c) a multi-layer stacked CNN (Sec.~\ref{sec:stacked_cnn}) that generates contextual vectors for the words in a given utterance, (d) three individual output layers (See.~\ref{sec:output_layers}) performing domain, intent and slot predictions based on the context vectors from Stacked CNN, and (e) two hierarchical link layers (See.\ref{sec:link_layers}) that transfer the posterior distribution of an upstream to a downstream task. We discuss each building block in detail in the following sections.

\subsection{Compositional CNN}
\label{sec:compositional_cnn}

The first layer of our network, named Compositional CNN (CompCNN), is used to create word representations using character embeddings. 
Let $\mathbb{C}$ be the set of characters and $d$ be the dimensionality of character embeddings. 
Let $\mathbf{V} \in \mathbb{R}^{d \times |\mathbb{C}|}$ be the matrix of character embeddings. Consider a word, $w$, that is made up of a sequence of $n$ characters $[c_1, c_2 ... c_n]$. For each character $c_i$ of word $w$, we obtain the corresponding character embedding $\mathbf{V}_{c_i}$ from the embedding matrix and then concatenate them to create a character-level representation $\mathbf{C}_{w} \in \mathbb{R}^{d \times n}$ of the word.

In order to create feature maps from character-level representations we apply a convolution operation on the window of $l$ characters and obtain a vector of features $\mathbf{u} \in \mathbb{R}^{n - l +1}$. Eq.~\ref{eq:char_to_word} describes this operation as follows:
\begin{equation}\label{eq:char_to_word}
\mathbf{u} = ReLU(\mathbf{C}_{w} * \mathbf{H} + {b}) 
\end{equation}
where $*$ is the convolution operation; $\mathbf{H} \in \mathbb{R}^{d \times l}$ is the convolution filter with width $l$; $b$ is the bias and $ReLU$ is the nonlinear activation function.

To obtain the most important n-gram captured by a given filter, we apply the max-over-time pooling as illustrated in Eq.~\ref{eq:max_pool_ngrams}. This operation returns the maximum valued feature $v$ in feature vector $\mathbf{u}$. This pooling scheme naturally facilitates the model to deal with the variable word lengths.
\begin{equation}\label{eq:max_pool_ngrams}
v = max \hspace{.1cm} \mathbf{u}
\end{equation}
The operations defined in Eq.~\ref{eq:char_to_word} and Eq.~\ref{eq:max_pool_ngrams} contributes only one n-gram feature per convolution filter. Therefore, we use multiple filters and obtain a feature vector of length equaling the number of applied convolution filters for the input word.
Further, we vary the filter width from 3 to 6 to extract n-gram features of corresponding sizes.

\subsection{Highway Network}
\label{sec:highhway_network}

We use a highway network \cite{srivastava2015training} which consists of a transform and a carry gate at two locations of our network; 
after the CompCNN layer and the stacked CNN layer. 
Highway networks not only enable us to apply non-linearity to the inputs but also facilitate training by carrying some of the input directly to the output.

Formally, highway networks are defined as follows:
\begin{equation}
\mathbf{y} = \mathbf{t} \odot f( \mathbf{W}_H \mathbf{x} + \mathbf{b}_H ) + \mathbf{(1-t)} \odot \mathbf{x}
\end{equation}
Where $\odot$ is element-wise multiplication; $\mathbf{x}$ and $\mathbf{y}$ are input and output vectors; 
$\mathbf{t} = \sigma( \mathbf{W}_T \mathbf{x} + \mathbf{b}_T ) $ and $\mathbf{(1-t)}$ denote the transform and the carry gate respectively; $f$ is the non-linearity which is $ReLU$ in this work and $\sigma$ is the sigmoid function; $\mathbf{W}_H, \mathbf{W}_T \in \mathbb{R}^{m \times m}$ are the weights of feature transform layer and transform gate, where m is the size of the input vector; $\mathbf{b}_H$, $\mathbf{b}_T$ are the corresponding bias vectors.

\subsection{Stacked CNN}
\label{sec:stacked_cnn}

In order to model interactions between the words of a sentence we employ a multi-layer CNN, which we call Stacked CNN, similar to 
the works of 
\cite{collobert2011natural, kim2014convolutional} 
where competitive results are presented for various NLP tasks.
We stack multiple convolution layers to increase the receptive field of the model which enables learning long-term dependencies among the input words.

\begin{figure}[t]
    \centering
    \includegraphics[width=.85\columnwidth]{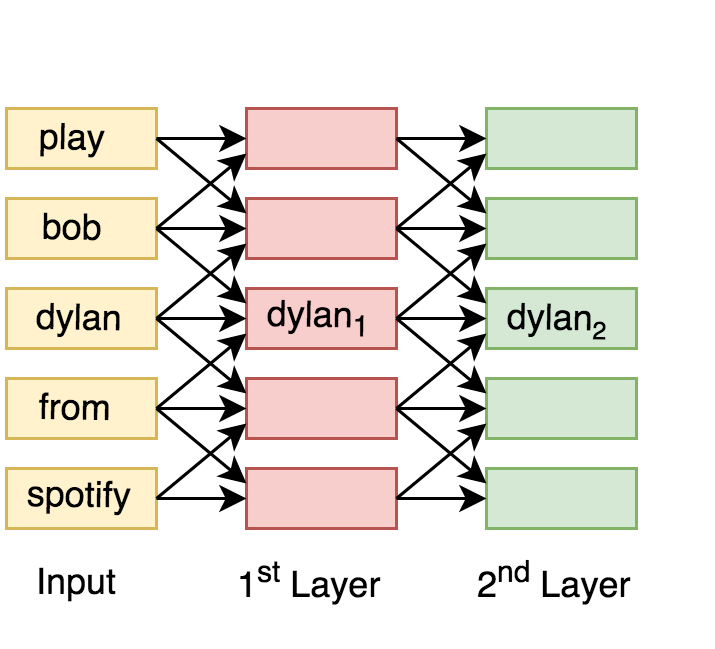}
    \caption{A sample stacked CNN with 2 CNN layers where each layer has a receptive field of 3 words. 
    As an example, while constructing the contextual vector of the word ``dylan'' in the first layer, $dylan_1$, only the words in the immediate neighbourhood is considered. On the other hand, in the second layer for the same word, $dylan_2$ is affected by the whole utterance.
    }
    \label{fig:stacked_cnn}
\end{figure}

Consider an input utterance $s$ as a sequence of words $w_1, w_2, ..., w_n$, where the vector representation $\mathbf{v}_{w_i}$ of word $w_i$ is the output of the highway network.
We concatenate these vectors to obtain a matrix representation $\mathbf{R}_{s} \in \mathbb{R}^{d \times n}$ of the input utterance $s$.
We pass $\mathbf{R}_s$ through the cascade of convolution layers to obtain the contextual vectors of the words of an utterance. We present a visual example for the process in Fig.~\ref{fig:stacked_cnn}.
The contextual vector of a word encodes information about the word itself and its adjacent words. A single layer of the Stacked CNN is defined as follows:
\begin{equation}\label{eq:stacked_cnn_features}
    \mathbf{z} = ReLU (\mathbf{R}_s * \mathbf{H} + \mathbf{b})
\end{equation}
Where $*$ is the convolution operation; $\mathbf{H} \in \mathbb{R}^{d \times l}$ is the convolution filter with width $l$ and $d$ is dimensionality of temporal vectors; $\mathbf{b}$ is the bias; $ReLU$ is the nonlinear activation; $\mathbf{z} \in \mathbb{R}^{n - l + 1}$ is the feature map capturing the local contexts.
In order to obtain fixed length vectors for each sentence, we pad each sentence to be of the same length.
As per Eq.~\ref{eq:stacked_cnn_features}, each convolution filter results in one feature for every context vector so we use multiple filters per CNN.

\subsection{Output Layers}
\label{sec:output_layers}

\begin{figure}
    \centering
    \includegraphics[width=0.47\textwidth]{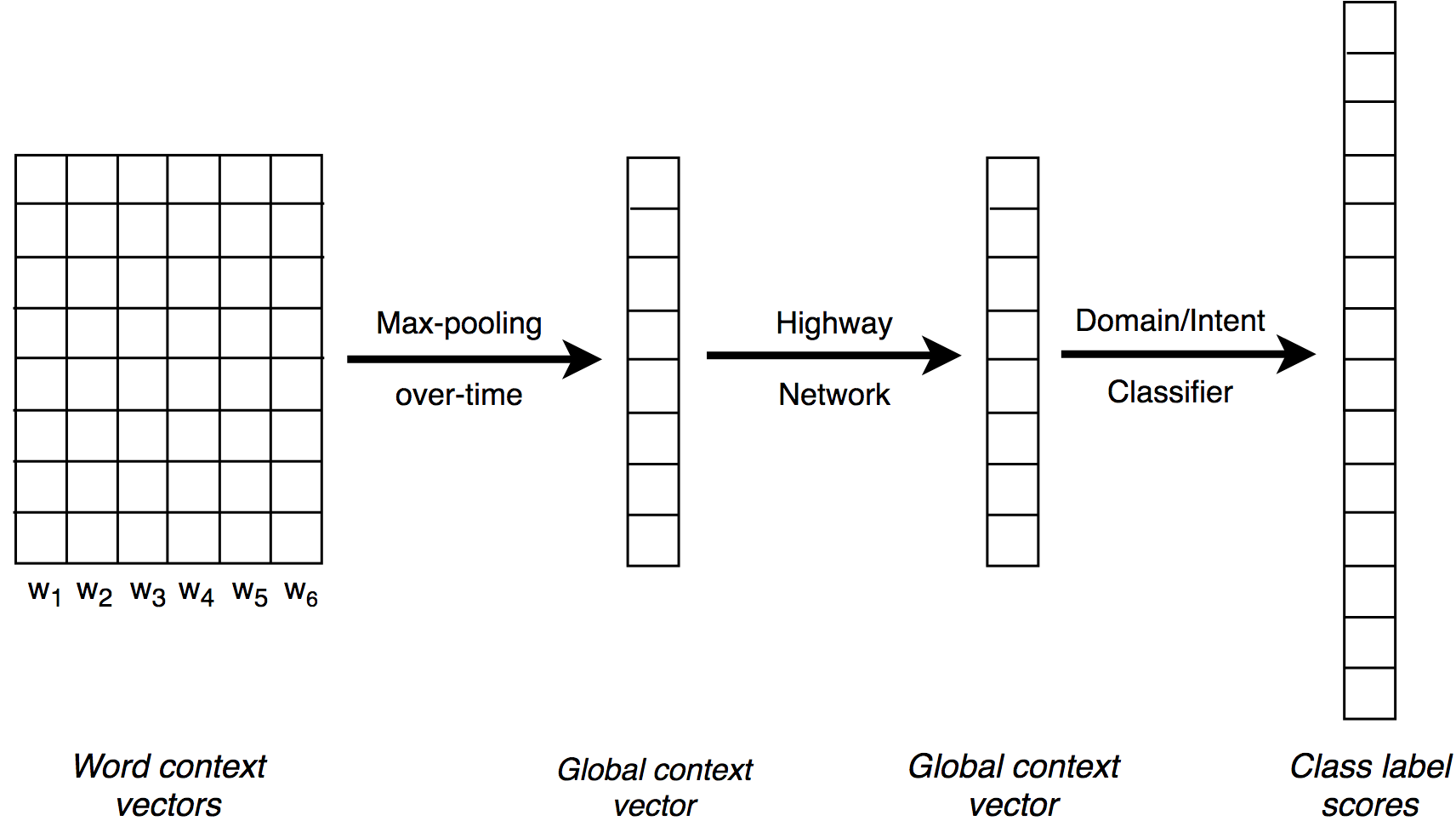}
    \caption{Processing of the word contexts in order to produce global-context classification results.}
    \label{fig:global_context}
\end{figure}

Our architecture models three classification tasks namely domain, intent and slot classification, where the first two tasks operate on the utterance level while the third task assigns a label to each word of the input utterance. We categorize domain and intent classification as global-context tasks and slot classification as a local-context task. We discuss these tasks below.

\subsubsection{Local-context Tasks}

The local-context task takes as input the context vector associated with the target word. 
That is to say, the output of the Stacked CNN that is associated with the  target word is used as input in order to find the label of the word.

During processing, we apply a highway layer to the input of the local-context task, and then use a fully connected layer followed by softmax to produce normalized classification scores.

\subsubsection{Global-context Tasks}

Processing of word context vectors for global-context tasks is visualized in Fig.~\ref{fig:global_context}. We obtain the vector representation of an utterance by applying a max-over-time pooling operation 
on the matrix of word context vectors such that its dimensionality equals to the number of feature maps produced by the last layer of Stacked CNN. Subsequently, we apply a highway network which is then input to a fully-connected layer followed by softmax to obtain task specific normalized classification scores. Since we have two global-context tasks, our architecture contains two such output layers.

\subsection{Hierarchical Link Layers}
\label{sec:link_layers}

Although the joint optimization of the NLU tasks elevates their individual performance, it is plausible to achieve further gains when each downstream task can access the posterior distributions of its upstream task. 
The motivation behind this claim is the human biology
where one can learn complex ideas more easily 
in the presence of some prior information about related simpler ideas. 

Therefore, our architecture includes two hierarchical link layers: (a) domain to intent and, (b) intent to slot where we transfer the information between the semantically hierarchical NLU tasks. 
Eq.~\ref{eq:di_link} and \ref{eq:di_sum} describe the domain-intent link layer and summation of its output $\mathbf{y}_{di}$ to the input $\mathbf{x}_{i}$ of the intent classifier to obtain the new input vector $\tilde{\mathbf{x}_{i}}$ respectively. Formally we define:
\begin{equation}
    \mathbf{y}_{di} = f ( \mathbf{W}_{di} \hspace{.1 cm}\mathbf{u}_{d} + \mathbf{b}_{di} ) \label{eq:di_link}
\end{equation}
\begin{equation}
    \tilde{\mathbf{x}_{i}} = \mathbf{y}_{di} + \mathbf{x}_{i} \label{eq:di_sum}
\end{equation}
where $\mathbf{u}_{d}$ is the input to the domain classifier softmax layer; 
$\mathbf{W}_{di}$ and $\mathbf{b}_{di}$ are the weights and bias of domain-intent link layer respectively; $f$ is a non-linear activation function.

We implement the intent-slot link layer similar to the domain-intent link layer as described in Eq.~\ref{eq:is_link}. Although addition of the link layer's output to the context vector of each word seems pretty reasonable, we use a gating mechanism to control the amount of link layer information added to the inputs of slot classifier. Such scheme ensures that the model furnish the intent information to some particular words only and avoid connecting common words to any single intent. For instance, consider ``play Mozart from Spotify'' where only ``Mozart'' (Artist name) and ``Spotify'' (Media service) are coupled strongly to ``Play music'' intent whereas words like ``play'' and ``from'' can occur in an utterance like ``play Hobbit from Audible'' which belongs to an intent from another domain. 
We implement this gating mechanism using a feedforward layer as defined in Eq.~\ref{eq:link_gate}, which transforms the input vector $\mathbf{x}_{i}$  of the slot classifier to determine their corresponding gating vectors. As described in Eq.~\ref{eq:is_sum}, these gating vectors are element-wise multiplied to the output vector $\mathbf{y}_{is}$ of the intent-slot link layer and later added to the input vector $\mathbf{x}_{i}$ to create a semantically rich input $\tilde{\mathbf{x}_{i}}$ for the slot classifier. Formally we define:
\begin{equation}
    \mathbf{y}_{is} = f ( \mathbf{W}_{is} \hspace{.1 cm}\mathbf{u}_{i} + \mathbf{b}_{is} ) \label{eq:is_link}
\end{equation}
\begin{equation}
    \mathbf{y}_{gate} = \sigma ( \mathbf{W}_{gate} \mathbf{x}_{i} + \mathbf{b}_{gate} ) \label{eq:link_gate}
\end{equation}
\begin{equation}
    \tilde{\mathbf{x}_{i}} = (\mathbf{y}_{gate} \odot \mathbf{y}_{is}) + \mathbf{x}_{i} \label{eq:is_sum}
\end{equation}
where $\mathbf{u}_{i}$ is the input to the intent classifier softmax layer;
$\odot$ denote elementwise multiplication; 
$\mathbf{W}_{gate}$ and $\mathbf{b}_{gate}$ are the weight matrix and bias vector of gating layer; $\sigma$ is the sigmoid non-linearity.

\section{Experiments}

\subsection{Data}\label{sec:data}

In this work, we use user requests in German to voice-controlled devices consisting of two datasets: (a) in-house collected (b) beta data. 
Beta data is collected from users in an open-ended environment and have a higher variance than the in-house collected data thus simulating the actual distribution of open-ended queries expected from the live users. 
We train our models using the in-house collected data and evaluate its performance on the beta data. 
This simulates the real life situation where an initial model needs to be provided to the beta user population before beta data becomes available. It is expected from the initial model to have a respectable accuracy to facilitate data collection and improve beta user experience.
Note that the results presented in this paper are not on production or live user traffic; rather, they reflect NLU bootstrap performance, which is limited to the data obtained through in-house data collection. Additional, significant improvements in accuracy can be obtained by training on larger volumes of data obtained once the system is available to larger sets of users. 
We train our models for various sizes of data to study the effect of the dataset size on the testing performance. The in-house and beta datasets comprise of 180,000 and 12,000,000 data samples respectively. Collectively, the dataset contains 21 domains, 194 intents and 163 slots. 


\subsection{Experimental Setup}
\label{sec:exp_setup}

In this work, we implement all the models using MXNet \cite{chen2015mxnet}. 
We represent each character by a 15-dimensional embedding vector. 
For CompCNN, we use four CNNs with kernel size 3, 4, 5  and 6 having 50, 75, 75 and 150 output channels respectively. 
Further, we use a two-layered CNN for the word-level Stacked CNN, where each layer contains 100 convolution filters with kernel size 3. 
To facilitate the residual connections in highway layers and avoid any dimensionality conflicts, we ensure that the inferred embedding vectors of words, their contexts, and sentence, must be of the same size, which is 100 in this work. We use Adam \cite{kingma2014adam} optimizer with learning rate $10^{-4}$ and Xavier \cite{glorot2010understanding} initialization for training our models.


\subsection{Effect of Character-level Modelling}
\label{sec:char_vs_word}

To study the benefits of character-level modeling, we compared the performance of character-level and word-level models.
In this experiment, we implement the word-level model by replacing the character embedding and CompCNN layers of our architecture with a word embedding layer where each column vector of the embedding matrix represents a word in the input vocabulary.

\begin{table}[tb!]
\centering
\resizebox{\columnwidth}{!}{
\begin{tabular}{lccccccc} 
 \hline
Training & \multicolumn{3}{c}{Character-level model} & &  \multicolumn{3}{c}{Word-level model} \\
\cline{2-4} \cline{6-8} 
size & Domain & Intent & Slot &  & Domain & Intent & Slot \\ 
\hline
10K & \textbf{79.59} & \textbf{73.75} & \textbf{69.55}  & & 74.65 & 66.36  & 59.47  \\

20K & \textbf{82.96} & \textbf{77.32} & \textbf{72.96}  & & 80.46 & 72.75  & 68.84 \\

40K & \textbf{86.25} & \textbf{81.09} & \textbf{77.46} & & 82.68 & 78.37  & 74.72 \\

80K & \textbf{87.72} & \textbf{83.47} & \textbf{80.23} & & 85.42 & 80.31  & 77.27 \\

160K & \textbf{88.26} & \textbf{84.42} & \textbf{81.58} & & 85.42 & 81.83  & 77.99 \\
 \hline
\end{tabular}
}
\caption[Performance comparison of character vs. word models]{F1 scores (\%) of domain, intent and slot classification tasks modelled by character-level and word-level models. We present the results for various sizes of training data and compare the performance. For each data size, the best result is highlighted in bold for each task.
}
\label{table:char_vs_word_beta}
\vspace{-0.1in}
\end{table}

Tab.~\ref{table:char_vs_word_beta} shows that the character-level model performs better than the word-level model for all classification tasks and all sizes of training data. 
For example, when using 10K training utterances, the character-level model achieves a relative improvement of 6.62\%, 11.14\% and 11.95\% in F1 scores of domain, intent and slot classification respectively compared to the word-level model. 

Since limited amounts of training data increases the number of out of vocabulary (OOV) words, the word-level model fails to capture the semantic information of an utterance reliably. On the other hand, character-level model handles the OOV words by modeling the words as a sequence of characters which alleviate the lack of training data to a certain extent. Moreover, the implicit parameter sharing between subword units facilitates the character-level model to learn semantically-rich word embeddings.

Another point of note is that using word embeddings significantly increases the number of free parameters of the model and thus a larger training set is required to be able to improve accuracy.

\subsection{Effect of Multi-task Learning}
\label{sec:single_vs_multi}

Multi-task Learning (MTL) facilitates implicit knowledge transfer among the jointly-modeled tasks via parameter sharing, which is especially beneficial in low-resource settings. Such joint modeling techniques generally improve the performance of each learning task of the model, especially the ones with higher complexity. In this experiment, we created single-task models by separating domain, intent and slot classification tasks and training on the task specific data. We then compared single-task models with the model where MTL is used. We present the experiment results where we compare single and multi task models in Tab.~\ref{table:single_vs_multi_beta}.

\begin{table}[tb!]
\centering
\resizebox{\columnwidth}{!}{
\begin{tabular}{lccccccc} 
 \hline
Training & \multicolumn{3}{c}{Multi-task model} & &  \multicolumn{3}{c}{Single-task models} \\
\cline{2-4} \cline{6-8} 
Size & Domain & Intent & Slot &  & Domain & Intent & Slot \\ 
\hline
10K & 79.59 & \textbf{73.75} & \textbf{69.55} & & \textbf{81.53} & 69.63 & 68.20  \\

20K & 82.96 & \textbf{77.32} & \textbf{72.96} & & \textbf{83.80} & 75.84 & 72.02 \\

40K & \textbf{86.25} & \textbf{81.09} & \textbf{77.46} & & 86.11 & 79.43 & 76.03 \\

80K & 87.72 & \textbf{83.47} & \textbf{80.23} & & \textbf{87.73} & 81.57 & 76.81 \\

160K & 88.26 & \textbf{84.42} & \textbf{81.58} & & \textbf{88.58} & 82.43 & 78.34 \\
 \hline
\end{tabular}
}
\caption[Performance of single-task vs. multi-task models]{F1 scores (\%) of domain, intent and slot classification tasks modelled by single-task vs. multi-task models. We present the results for various sizes of training data and compare the performance. For each data size, the best result is highlighted in bold for each task.}
\label{table:single_vs_multi_beta}
\vspace{-0.1in}
\end{table}

For the domain classification, we observe that the single-task model achieves better performance most of the time. 
The domain classification task is the most basic task in our NLU system, and experiments show that a dedicated domain classifier might be enough to get a decent performance. This is the only task where we observe little to no benefit of using MTL.

On the other hand, the multi-task model performs better for intent classification for all sizes of training data. Due to a large number of intent labels, the complexity of the intent classification is much higher than the domain classification. Additionally, some intents like ``Play'' intent that occur in multiple domains like ``Books'' and ``Music'', makes the intent classifier prone to ambiguity. Since MTL enables information sharing, the input vectors of intent classifier also includes the information about the domain which help the intent classifier focus more on the semantics of ``Play'' command rather than the domain related information, thus improving its classification accuracy.

Slot classification performs fine-grained semantic analysis of the input utterance by assigning a named entity to every word. Since it is a local-context task, a single-task model infers the slot of a word merely based on its neighboring words. On the other hand, the multi-task model takes advantage of the supplementary information provided by the domain and intent of the input utterance which helps to infer the correct slots. In utterances like ``Play Beatles from Spotify'' and ``Play Avengers on Netflix'', accurate predictions of named entities i.e., ``Band name'' for ``Beatles'' and ``Movie name'' for ``Avengers'', are crucial for executing the user command correctly. Similar to intent classification, the F1 scores of multi-task model are better compared to the single-task model for the slot classification for all data sizes. 

\subsection{Effect of Pre-trained Embeddings}
\label{sec:pre_trained}

Since pre-trained word embeddings are trained on a large corpus of text like Wikipedia, they cover a significant amount of structured language which enables them to encode highly robust semantic information for each word. 
For transferring the semantic knowledge of the pre-trained word embeddings to our character-level architecture, we train only the first three layers of our model using L2 loss such that the word-embeddings produced by the network are similar to the pre-trained embeddings.
Subsequently, we use these trained parameters to initialize the first three layers of the model while randomly initializing the other layers. We use German word-embeddings obtained from fastText \cite{grave2018learning} and call the model where the first three layers are pre-trained as the FastText-init model.

\begin{table}[t!]
\centering
\resizebox{\columnwidth}{!}{
\begin{tabular}{lccccccc} 
 \hline
Training & \multicolumn{3}{c}{FastText-init model} & &  \multicolumn{3}{c}{Randomly initialized model} \\
\cline{2-4} \cline{6-8} 
Size & Domain & Intent & Slot &  & Domain & Intent & Slot \\ 
\hline
10K  & \textbf{81.59} & \textbf{75.81} & \textbf{72.01} & & 79.59 & 73.75 & 69.55 \\
20K  & \textbf{84.88} & \textbf{79.86} & \textbf{76.16} & & 82.96 & 77.32 & 72.96 \\
40K  & \textbf{87.41} & \textbf{82.59} & \textbf{79.19} & & 86.25 & 81.09 & 77.46 \\
80K  & \textbf{88.37} & \textbf{84.31} & \textbf{81.09} & & 87.72 & 83.47 & 80.23 \\
160K & \textbf{88.93} & \textbf{85.12} & \textbf{82.51} & & 88.26 & 84.42 & 81.58 \\
 \hline
\end{tabular}
}
\caption[Performance of random vs. fastText init models on beta data]{F1 scores (\%) of domain, intent and slot classification tasks modelled by random vs. fastText-init models. We present the results for various sizes of training data and compare the performance. For each data size, the best result is highlighted in bold for each task. }
\label{table:random_vs_fast_beta}
\vspace{-0.1in}
\end{table}

We present the comparison of the random vs FastText-init models in Tab.~\ref{table:random_vs_fast_beta}. 
The FastText-init model performs better than the randomly initialized model for all NLU tasks and on all sizes of training data. 
We observe the biggest relative gains in the slot classification task which relies on robust word-embeddings, and benefits the most from pre-training even when trained with large amounts of data. 
This also shows that using character embeddings does not prevent us from taking advantage of the rich literature on word embeddings and using existing word embeddings to improve model performance.

\subsection{Effect of Hierarchical Link Layers}
\label{sec:link_layers_exp}

In this section, we discuss the training strategy we employ when using hierarchical link layers in our proposed multi-task architecture. Recall that the hierarchical link layers are the information pathways which transfer the output distribution from an upstream task's classifier to the input of its immediate downstream task, as defined in the semantic hierarchy of the NLU system. To compare the effect of these link layers, we train two separate models, one with hierarchical link layers called with-link model and another without it called no-link model. 

In our experiments, we observed that by naively optimizing a randomly initialized with-link model, the performance of domain and intent tasks deteriorates. 
We believe this is due to the fact that each task is trying to optimize its own output and the tasks working antagonistically blocking training progress. Therefore, we overcome this problem by training the with-link model in two steps. 
In the first step, we start training by disabling the link layers. This step not only partially optimizes the classifier of each task but also initializes the shared layers of our model and tunes each task to work together rather than against each other.
In the following step, we enable the hierarchical link layers and continue training using this partly optimized model, ensuring stable training dynamics, which balances the performance of each NLU task.

\begin{figure}[b!]
    \centering
    \includegraphics[width=0.45\textwidth]{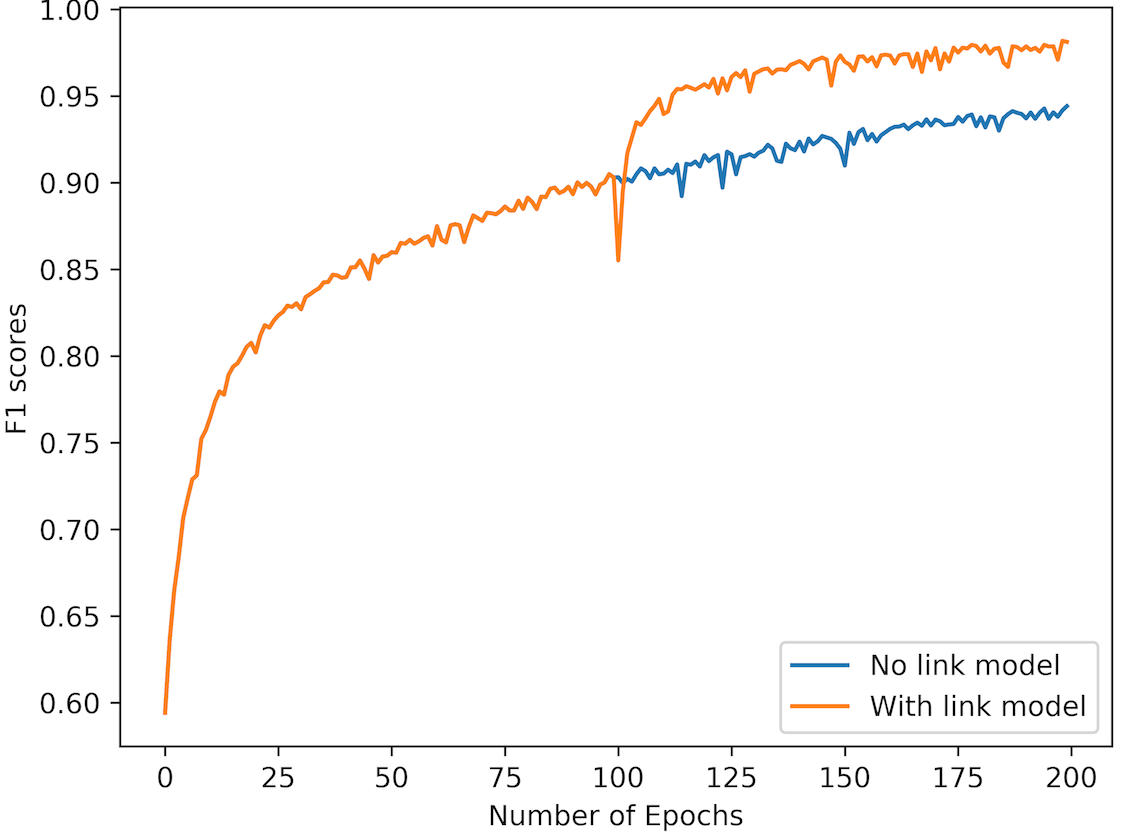}
    \caption{Validation accuracy during training of the with-links model for slot classification. Until the first 100 epochs the model is trained without links where we introduce the hierarchical links. While the slot classification accuracy drops in the first epoch, we observe a jump in the validation accuracy in the following epochs. }
    \label{fig:training_slot_links}
\end{figure}

We visualize the slot classification scores of the model on the validation set during training in Fig.~\ref{fig:training_slot_links}. In the first 100 epochs we disable the links between the classification tasks and train the remaining layers of our model. After that point, we enable the domain-intent and intent-slot links. As presented in the figure, while this causes a small drop in the validation accuracy at the first epoch the links are introduced, the model quickly recovers and achieves a significant jump in the slot classification validation accuracy in the following epochs.

\section{Conclusion}
\label{sec:conclusion}

In this work, we presented an end-to-end neural architecture for the joint modeling of the NLU tasks namely domain, intent and slot classification. 

For modeling extensive vocabularies of morphologically-rich languages like German and French, we used character-level modeling to compose word embeddings which enabled parameter sharing among words with similar subword units. 
In our architecture we did not rely on a morphological analyzer or disambiguator but rather adopted a data driven approach to extract useful information from the subwords. 
We showed that, the proposed character-level model performs notably better than the corresponding word-level models for all NLU tasks, and for all data sizes. 

We overcame the problem of error propagation from an upstream to a downstream task by joint modeling of all NLU tasks via MTL. We followed hard-parameter sharing approach to introduce shared hidden layers in our architecture, which permits information sharing among NLU tasks to learn domain-invariant features. We showed that the multi-task models obtain higher F1 scores than the single-task models for all NLU tasks. 
The performance gains were not just limited to models trained with small datasets, thereby indicating that information sharing among related tasks is extremely valuable irrespective of available training data. 

Additionally, we utilized pre-trained word embeddings to initialize our model for analyzing the impact of external knowledge sources on the performance of the model. We confirmed that the models initialized with pre-trained embeddings performed better than the randomly initialized models for all classification tasks. Further, we presented a simple technique to transfer the encoded semantic knowledge of pre-trained embeddings to the compositional layers of our character-level model. 

We conclude the proposed neural architecture is an efficient approach for bootstrapping the NLU module with minimal training data. 
The experiments of this work show that the character-level models combined with multi-task learning considerably improves the overall performance of the NLU module thus saving time, cost and effort to extend the current systems to new languages.

\bibliography{acl2019}
\bibliographystyle{acl_natbib}

\end{document}